\documentclass{article}
\usepackage{ijcai19}

\usepackage{times}
\usepackage{soul}
\usepackage{url}
\usepackage[hidelinks]{hyperref}
\usepackage[utf8]{inputenc}
\usepackage[small]{caption}
\usepackage{graphicx}
\usepackage{amsmath}
\usepackage{booktabs}
\usepackage{float}
\usepackage[minipage]{}

\title{Resource-aware Elastic Swap Random Forest \\ for Evolving Data Streams}

\author{
Diego Marr\'on$^{1,2}$\footnote{Contact Author}\and
Eduard Ayguad\'e$^{1,2}$\and
Jos\'e R. Herrero$^2$\and
Albert Bifet$^3$\\
\affiliations
$^1$ Computer Sciences Department, Barcelona Supercomputing Center, Barcelona, Spain\\
$^2$ Computer Architecture Department, Universitat Polit\`ecnica de Catalunya, Barcelona, Spain \\
$^3$ LTCI, T\'el\'ecom ParisTech, Universit\'e Paris-Saclay, 75013 Paris, France\\
\emails
\{diego.marron,eduard.ayguade\}@bsc.es,
josepr@ac.upc.edu,
albert.bifet@telecom-paristech.fr
}

\def\erf{\textsc{Elastic Random Forest}\xspace}
\def\srf{\textsc{Swap Random Forest}\xspace}
\def\esrf{\textsc{Elastic Swap Random Forest}\xspace}
\def\arf{\textsc{Adaptive Random Forest}\xspace}

\def\ewma{\textrm{EWMA}}

\usepackage{graphicx}
\usepackage{xspace} 
\usepackage{algorithm}
\usepackage{algpseudocode}

\begin{document}

\maketitle

\begin{abstract}
Continual learning based on data stream mining deals with ubiquitous sources of Big Data arriving at high-velocity and in real-time. Adaptive Random Forest ({\em ARF}) is a popular ensemble method used for continual learning due to its simplicity in combining adaptive leveraging bagging with fast random Hoeffding trees. While the default ARF size provides competitive accuracy, it is usually over-provisioned resulting in the use of additional classifiers that only contribute to increasing CPU and memory consumption with marginal impact in the overall accuracy. This paper presents Elastic Swap Random Forest ({\em ESRF}), a method for reducing the number of trees in the ARF ensemble while providing similar accuracy. {\em ESRF} extends {\em ARF} with two orthogonal components: 1) a swap component that splits learners into two sets based on their accuracy (only classifiers with the highest accuracy are used to make predictions); and 2) an elastic component for dynamically increasing or decreasing the number of classifiers in the ensemble. The experimental evaluation of {\em ESRF} and comparison with the original {\em ARF} shows how the two new components contribute to reducing the number of classifiers up to one third while providing almost the same accuracy, resulting in speed-ups in terms of per-sample execution time close to 3x.

\end{abstract}

\section{Introduction}

Nowadays, ubiquitous sources of Big Data are generating an unprecedented amount of dynamic Big Data streams {\em (Volume)}, at high speed {\em (Velocity)}, with newer data rapidly superseding older data {\em (Volatility)}. For example, the \textsc{Internet of Things} is the largest network of
sensors and actuators connected by networks to computing systems. This includes sensors across a huge range of settings, for industrial process control, finance, health analytics, home automation and autonomous cars, and often interconnected across domains, monitoring and functioning in people, objects and machines in real-time. 
Extracting knowledge on-the-fly from these Big Data streams, requires fast incremental algorithms that are able to deal with potentially infinite streams.  Also, algorithms should be adaptive, showing  the capability of adapting to the evolution of data distributions over time, since changes in them, can cause predictions to become less accurate {\em (Concept Drift)} .

Ensemble learners are the preferred method for processing evolving data streams due to their better classification performance over single models. Among ensemble learners, \arf ({\em ARF})~\cite{Gomes2017} is considered the state-of-the-art ensemble for classifying evolving data streams in the MOA~\cite{MOA} open source infrastructure. {\em ARF} requires few hyperparameters to run and adapts to concept drifting by combining adaptive leveraging bagging with fast random {\em Hoeffding} decision trees~\cite{DomingosH00}. {\em Hoeffding} trees are used for mining non-evolving data streams doing a single pass over the data; they are based on the theoretical guarantees of the {\em Hoeffding bound} to discover how many input samples are needed to decide when to grow the tree.

{\em ARF} uses by default 100 random decision trees, a number that is decided independently of the characteristics of the data stream. In this paper, we argue that this number of learners in {\em ARF} is not optimal and that we can get similar results using a smaller number of learners. For this, we introduce a  new adaptive methodology to automatically decide the number of learners to be used in incremental models.

We present a new ensemble method that extends the originally proposed {\em ARF} in two orthogonal directions. On one side, \esrf (\textit{ESRF}) splits the learners in two groups: a forefront group that contains only the learners used to do predictions, and a second  candidate group that contains learners trained in the background and not used to make predictions. At any time, a learner in the forefront group can be replaced by a candidate learner if this swapping operation improves the overall ensemble accuracy. On the other side, \textit{ESRF} extends {\em ARF} by dynamically increasing/decreasing the number of learners in the ensemble, in particular the number of learners in the forefront set. 



The paper is organised as follows: Section \ref{sec::related_work} presents the necessary background. Section \ref{sec::esrf} describes the proposed {\em ESRF} ensemble. The experimental evaluation and comparison with {\em ARF} is presented in Section \ref{sec::eval}.  Finally, Section \ref{sec::conclusion} concludes the paper and outlines some future work.


\section{Background and motivation}\label{sec::related_work}

The \textsc{Hoeffding Tree} ({\em HT})~\cite{DomingosH00} is a very fast incremental decision tree learner for large data streams that assumes that the data distribution is not changing over time. {\em HT} grows incrementally based on the theoretical guarantees of the {\em Hoeffding bound}; a node is expanded as soon as there is sufficient statistical evidence that an optimal splitting feature exists. The model learned by {\em HT} is asymptotically nearly identical to the one built by a non-incremental (batch) learner, if the number of training instances is large enough. 

\arf ({\em ARF}) is an adaptive ensemble algorithm for data stream mining that extends the original {\em Random Forest} ({\em RF})~\cite{Breiman2001} to the more challenging data stream setting. {\em RF} is an ensemble method that combines bagging with random decision trees. {\em ARF} incrementally works by doing a single pass over the data and handles concept drifting~\cite{Gama:2014:SCD:2597757.2523813}. 
A drift detector is associated to each tree in the ensemble. The algorithm uses two thresholds: a first permissive threshold to signal a drift warning;  and a second threshold to confirm the drift detection. As soon as a random decision tree signals a drift warning, {\em ARF} starts building a background learner for that tree that is not used for doing the ensemble prediction; this background learner can replace the learner if the warning escalates to a drift. 

\begin{figure}[h]
\includegraphics[scale=0.5]{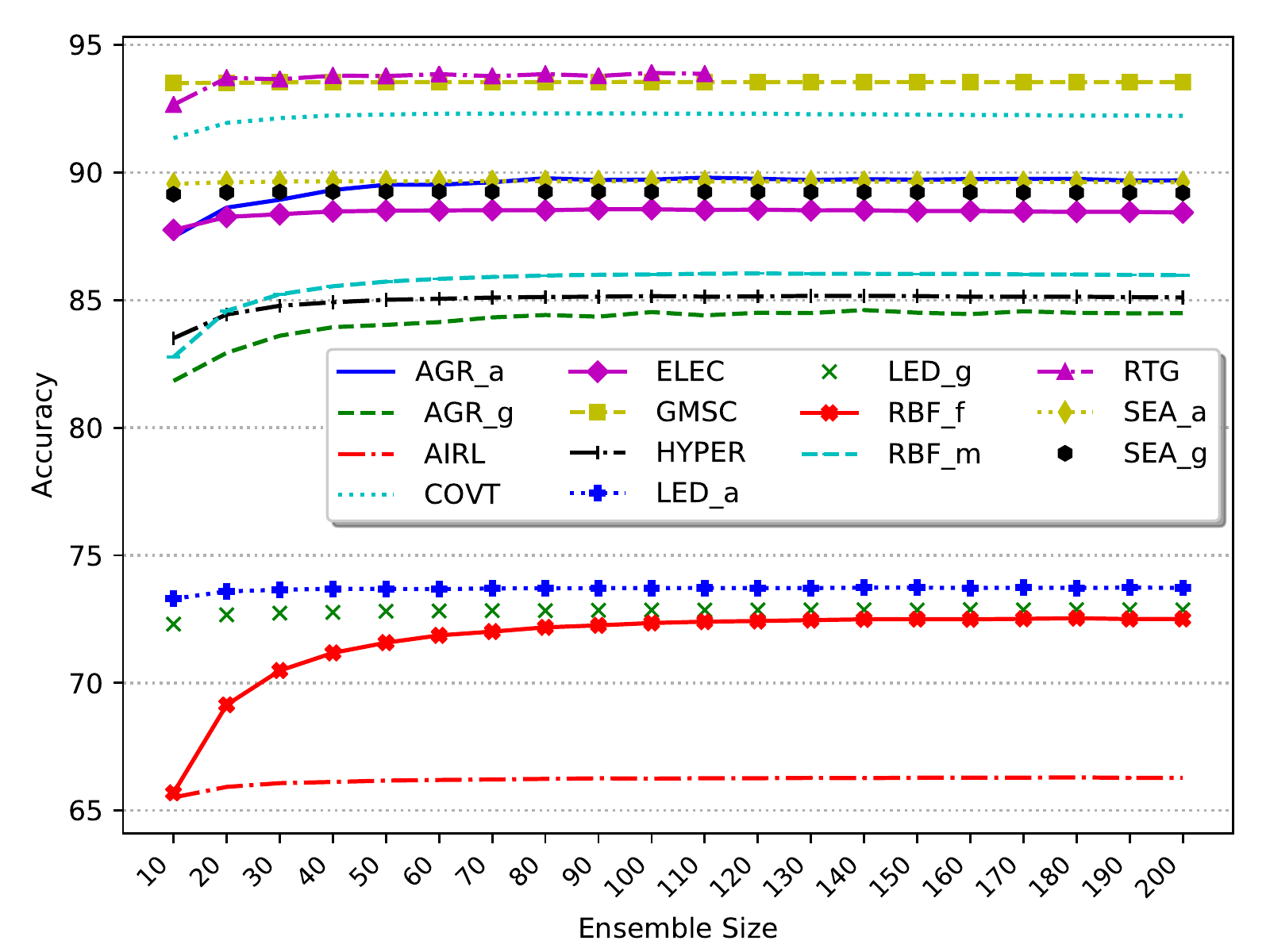}
\vspace{-0.25cm}
\caption{ARF accuracy evolution with ensemble size.}
\label{fig::arf_accuracy_evolution}
\end{figure}

By default, the reference implementation for {\em ARF} in the {\em Massive Online Analysis (MOA)} framework uses 100 random trees (ARF100). While this configuration provides competitive accuracy we have observed that accuracy can converge for sizes lower than 100 random trees, which is the case for almost all datasets used in this paper as shown in Figure \ref{fig::arf_accuracy_evolution}. The convergence point is different for each dataset, but once the accuracy converges, adding extra trees only increases computational and memory costs with marginal impact in the accuracy. For example, the RBF\_f dataset requires 70 learners in the ensemble to achieve a percentage difference in terms of accuracy (with respect to ARF100 ) in the second decimal place; however, the COVT dataset achieves the same difference in accuracy with only 30 learners.

In terms of memory, in the worst case scenario {\em ARF} allocates twice the number of active learners in the ensemble; however, background learners are simpler since they do not need to keep any drift detection data structure. In terms of execution time, each active and background tree needs to be traversed for each new arriving instance. With the aim of reducing the computational and memory requirements of the ensemble, enabling either high--throughput implementations or their deployment in resource--constrained devices, this paper proposes the \esrf algorithm described in the next section. 


\section{\esrf}\label{sec::esrf}

\esrf ({\em ESRF}) is a fast streaming random forest based method that adapts its size in an elastic way, to be consistent with the current distribution of the data. {\em ESRF} also includes a swap component that maintains two pools of classifiers to decide which ones are actually used for better prediction making. Although the elastic and swap components in {\em ESRF} are independent, they are presented together in Algorithm \ref{alg::esrf}. 

\begin{algorithm}[!b]
\caption{\esrf}
\label{alg::esrf}
\begin{algorithmic}[1]
\Require
\Statex $X$: input sample
\Statex $z$: label for input sample
\Statex $FS$: Set of forefront learners
\Statex $CS$: Set of candidate learners
\Statex $r$: Resize factor
\Statex $GS$: Set of $r$ random trees 
\Statex $f_{min}$: Classifier from $FS$ with lowest accuracy
\Statex $c_{max}$: Classifier from $CS$ with highest accuracy

\Function{TrainOnInstance}{$X$} 
\State \textsc{ResizeEnsemble}($X$, $z$)
\State \textsc{TrainAllClassifiers}($X$)
\State find $c_{max}$ and $f_{min}$
\State swap classifiers if $c_{max}$ is more accurate than  $f_{min}$ 
\EndFunction

\Function{TrainAllClassifiers}{$x$}
\For {each classifier $c \in FS \cup CS \cup GS$}
\State $w$ $\leftarrow$ Poisson($\lambda=6$) 
\State Set the weight of $x$ to $w$
\State Train classifier $c$ using $x$
\EndFor
\EndFunction

\Function{PredictOnInstance}{$X$} 
\State return \textsc{PredictLabel}($X$, $FS$)
\EndFunction

\Function{PredictLabel}{$x$, $S$}
\For {each classifier $c \in S$}
\State  $\hat{y_c} \leftarrow$ prediction for $x$ using $c$
\EndFor
\State $\hat{y} \leftarrow$ combination of predictions $\hat{y_c}$ 
\State return $\hat{y}$
\EndFunction
\end{algorithmic}
\end{algorithm}


The swap component in {\em ESRF} divides the classifiers into two groups: the foreground (or active) learners and the background (or candidate) learners. The {\em Forefront Set} ($FS$) contains those classifiers with higher accuracy that are used for predicting; the {\em Candidates Set} ($CS$) contains those classifiers that are trained but nor used for prediction since they accuracy is low compared to those on the $FS$. For each arriving instance $X$, the prediction is done just using the learners in $FS$ (lines 14--16). During training (lines 1--6), as done in {\em ARF}, all classifiers are trained simulating bagging by weighting the instance according to a $Poisson(\lambda=6)$. After that (lines 4--5), {\em ESRF} swaps the worst classifier in $FS$ (i.e. the one with the lowest accuracy $f_{min}$) with the best in $CS$ (i.e. the one with the highest accuracy $c_{max}$) when the later becomes more accurate. With the swap component the number of learners required in the ensemble is $|FS| + |CS|$; all the learners in these two sets are trained for each arriving instance.  

The elastic component in {\em ESRF} dynamically determines the size of the $FS$. This elastic component is not tied to {\em ESRF} and it could be implemented in any ensemble method such as the original {\em ARF}. For each arriving instance, the algorithm checks if $FS$ needs to be resized (line 2), adding or removing $r$ new learners. In case of growing, the new $r$ learners could be taken from the $CS$ set; however, and in order to add more diversity in the ensemble, the elastic component introduces a third set of learners, the {\em Grown Set} ($GS$) with $r$ learners, which also nees to be trained on each arriving instance and is reset on every resizing operation.

Algorithm \ref{alg::esrf:check_if_resize} details how the elastic component in {\em ESRF} works. 
For each arriving instance, {\em ESRF} simulates having three ensembles of random trees (lines 2--4): 
\begin{itemize}
    \item the default one (i.e., the current ensemble with $|FS|$ learners in $FS$); 
    \item the shrunk ensemble containing only the $|FS|-r$ learners with higher accuracies in $FS$;
    \item and the grown ensemble containing the learners in $FS$ FS plus the $r$ extra learners in the Grow set $GS$ (i.e. in total $|FS|+r$ learners).
\end{itemize}

\noindent Line 5 decides whether resizing is needed, based on the performance of each of these three ensembles. {\em ESRF} may decide to keep the current configuration for the ensemble or to apply a resize operation if either the shrunk or grown ensemble improve the default ensemble performance. In this case, the $r$ learners in $GS$ are added to $FS$ in case growing is decided (lines 6--9) or the $r$ classifiers in $FS$ with lowest accuracy are removed in case shrinking is decided (lines 10--13). Observe that the proposed elastic component does not use the candidates set $CS$, although this is an alternative option that is not considered due to the limit in the number of pages. 

\begin{algorithm}[h]
\caption{Check resize and update size for {\em ESRF}}
\label{alg::esrf:check_if_resize}
\begin{algorithmic}[1]
\Require

\Statex $r$: Resize factor
\Statex $FS$: Set of forefront learners
\Statex $FS_{min}$: Set of $r$ learners from $FS$ with lower accuracy 
\Statex $GS$: Set of $r$ random trees 
\Statex $SRK$: Shrunk ensemble $FS \setminus FS_{min}$
\Statex $GRN$: Grown ensemble $FS \cup GS$

\Statex $T_g$: grow threshold
\Statex $T_s$: shrink threshold

\Function{ResizeEnsemble}{$x$, $z$} 
\State $\hat{y_s} \leftarrow$ \textsc{PredictLabel}($x$, $SRK$)
\State $\hat{y_d} \leftarrow$ \textsc{PredictLabel}($x$, $FS$)
\State $\hat{y_g} \leftarrow$ \textsc{PredictLabel}($x$, $GRN$)
\State Operation $\leftarrow$ \Call{CheckIfResize}{$z$, $\hat{y_s}$, $\hat{y_d}$, $\hat{y_g}$}
\If{Operation==GROW}
\State $FS = FS \cup GS$
\State Start new $GS$ with $r$ new trees
\EndIf
\If{Operation==SHRINK} 
\State $FS = FS \setminus FS_{min}$
\State Start new $GS$ with $r$ new trees
\EndIf
\EndFunction

\Function{CheckIfResize}{$z$, $\hat{y_s}$, $\hat{y_d}$, $\hat{y_g}$}
\State Update $\ewma_{shrunk}$ using $\hat{y_s}$ and $z$
\State Update $\ewma_{default}$ using $\hat{y_d}$ and $z$
\State Update $\ewma_{grow}$ using $\hat{y_g}$ and $z$
\State $\Delta_{shrink} = \ewma_{shrunk} - \ewma_{default}$
\State $\Delta_{grow} = \ewma_{grown} - \ewma_{default}$
         
\If{ $\Delta_{grow} > \Delta_{shrink}$ and { $\Delta_{grow} > T_g$}  } 
\State 	return GROW;
\EndIf
\If{ $\Delta_{shrink} > \Delta_{grow}$ and $\Delta_{shrink} > T_s$}
\State	return SHRINK
\EndIf
\EndFunction

\end{algorithmic}
\end{algorithm}

The performance for an ensemble is computed using the exponential weighted moving average (EWMA) of its accuracy. EWMA gives larger weight to recent data, and a smaller weight to the older one. The weighting factor decreases exponentially but never reaches zero. It is calculated using this formula:

$$
\ewma_i=\ewma_{i-1} + \alpha * (S_i - \ewma_{i-1})
$$
where $S_i$ is the current value being added, $\alpha$ is the weighting factor defined as $\alpha = exp({1}/{W})$, where $W$ is a fixed time window. In {\em ESRF}, $W=2000$ and $S_i \in \{0,1\}$, $1$ for a label predicted correctly and $0$ otherwise. EWMA allows {\em ESRF} to keep track of each ensemble accuracy without being influenced too much by past prediction results.  

The necessary logics to decide whether the ensemble should be resized or not are detailed in function \textsc{CheckIfResize} in Algorithm \ref{alg::esrf:check_if_resize}. 
First, it updates each ensemble EWMA (lines 16--18); then compares the EWMA estimation of the default ensemble with the other two ensembles to compute the differences $\Delta_{shrink}$ and $\Delta_{grown}$ (line 19--20).  If $\Delta_{grown}$ is larger than $\Delta_{shrink}$, and $\Delta_{grown}$ is above a threshold, then a grow operation is triggered (lines 21--23). The shrink operation is decided in lines 24--26 and works similarly. 
In case $\Delta_{grown} = \Delta_{shrink}$, {\em ESRF}  favours growing against shrinking by comparing $\Delta_{grown}$ first.



As mentioned above, only classifiers in $FS$ are used to make the ensemble prediction (lines 14--16 in Algorithm \ref{alg::esrf}). {\em ESRF} implements the same weighting voting policy for instances as in {\em ARF}: each classifier has an associated weight that is computed as the number of correctly classified instances divided by the total number of instances since last reset (due to concept drift), reflecting the classifier performance on the current concept. To cope with evolving data streams, a drift detection algorithm is used with each learner of the ensemble algorithm described above (not shown in algorithm \ref{alg::esrf}). {\em ESRF} resets a tree as soon as it detects concept drifting. This is much simpler than the drift detection algorithm in {\em ARF} since there is a single threshold and no background learners are created when drift is detected. 

\section{Experimental Evaluation}\label{sec::eval}

\esrf ({\em ESRF}) has been implemented in the MOA (Massive Online Analysis) framework \cite{MOA}, an open source software environment for data stream mining, that implements a large number of data stream learning methods, including \arf ({\em ARF}). MOA is written in Java, making easier the prototyping and evaluation of novel online learning proposals and their comparison with state-of-the-art algorithms. 

Fourteen datasets have been used for the evaluation of {\em ESRF} and comparison with {\em ARF}. Ten of these datasets are synthetically generated using well known data generators: LED \cite{LEDGen}, SEA \cite{SEAGen}, Agrawal \cite{AGRAWALGen}, Random Tree Generator (RTG) \cite{DomingosH00}, Radial Basis Function (RBF) \cite{HYPERGEN} and Hyperplane \cite{HYPERGEN}). The datasets generated with the first three generators simulate both abrupt and gradual drifts. The dataset generated with the RTG generator does not simulate drift. The three datasets generated with RBF and Hyperplane simulate incremental (moderate or fast) drifts. In addition to these synthetic datasets, four real--world datasets that are widely used in the literature to evaluate data stream classification are also used: Airlines (AIRL) \cite{DS_Airlines}, Electricity \cite{DS_ELEC}, Forest Covertype \cite{DS_COV} and Give Me Some Credit (GMSC) \cite{DS_GMSC}. Table \ref{tab::datasets} presents an overview of these data sets, including number of samples, attributes and labels, as well as drift type.

\begin{table}[h]
\footnotesize
\caption{Synthetic (top) and real-world (bottom) datasets used for performance evaluation and comparison. Synthetic datasets drift type: A (abrupt), G (gradual) I.F (incremental fast), I.M (incremental moderate), N (None)}
\label{tab::datasets}
\centering
\begin{tabular}{|l|rcc|l|l|}
\hline
Dataset & Samples  &  Attrs & Labels & Generator & Drift \\
\hline
AGR\_a & 1,000,000 & 9 & 2 & Agrawal & A\\
AGR\_g & 1,000,000 & 9 & 2 & Agrawal & G\\
HYPER  & 1,000,000 & 10 & 2 & Hyperplane & I.F\\
LED\_a & 1,000,000 & 24 & 10 & LED Drift & A \\
LED\_g & 1,000,000 & 24 & 10 & LED Drift & G\\
RBF\_m & 1,000,000 & 10 & 5 & RBF & I.M\\
RBF\_f & 1,000,000 & 10 & 5 & RBF & I.F\\
RTG    & 1,000,000 & 10 & 2 & RTG & N\\
SEA\_a & 1,000,000 & 3 & 2 & SEA & A\\
SEA\_g & 1,000,000 & 3 & 2 & SEA & G\\
\hline
AIRL  & 539,383 & 7 & 2 &- &  -\\
COVT  & 581,012 & 54 & 7 & - & -\\
ELEC  & 45,312  & 8 & 2 & - &  -\\
GMSC  & 150,000 & 11 & 2 &- & -\\
\hline
\end{tabular}
\end{table}

The hardware platform that has been used to conduct the performance analysis is an Intel(R) Xeon(R) Platinum 8160 CPU running at 2.10GHz (24 cores, 48 threads), 96GB of RAM, SUSE Linux Enterprise Server 12 SP2 (kernel 4.4.120-92.70-default) and openJDK 64bits 1.8.0\_161.

We gradually evaluate the impact of the two components in the proposed {\em ESRF}. First we evaluate the performance of the swap component and then we evaluate the impact of adding the elastic component, always comparing against the baseline {\em ARF} ensemble with 100 learners (ARF100).  Two parameters in {\em ESRF} are fixed in the evaluation presented in this paper: $|CS| = 10$ and $r=|GS|=1$. Regarding the number of learners in {\em CS}, we have observed that it does not have a major impact in terms of accuracy for values larger than 10; in order to make a fair comparison in terms of resources when comparing with {\em ARF}, we set the value to 10, which coincides with the average number of background learners that are required by the drift mechanism in {\em ARF} for the datasets used in this paper. Regarding the resize factor in the elastic component, we have evaluated values of 1, 2 and 5, but not included the results due to lack of space since it does not have a significant impact in terms of accuracy. The parameters that are evaluated are: $|FS|$ (with a minimum value of 15 and maximum limited to $|FS|+|CS|+|GS|=100$) and the two thresholds that decide resizing ($T_g$ and $T_s$). The rest of hyper-parameters, which are common for both ESRF and ARF100, are set to their default values in MOA.

The evaluation methodology that has been used to conduct all the experiments reported is streaming prequential 10-fold cross-validation ~\cite{BifetMRHP15}. 

\subsection{\srf}

This subsection individually evaluates the swap component in {\em ESRF} in terms of accuracy and ensemble size, using ARF100 as a reference.
We will refer to this {\em ESRF} ensemble configuration as \srf ({\em SRF}). 

\begin{figure}[h]
\includegraphics[scale=0.5]{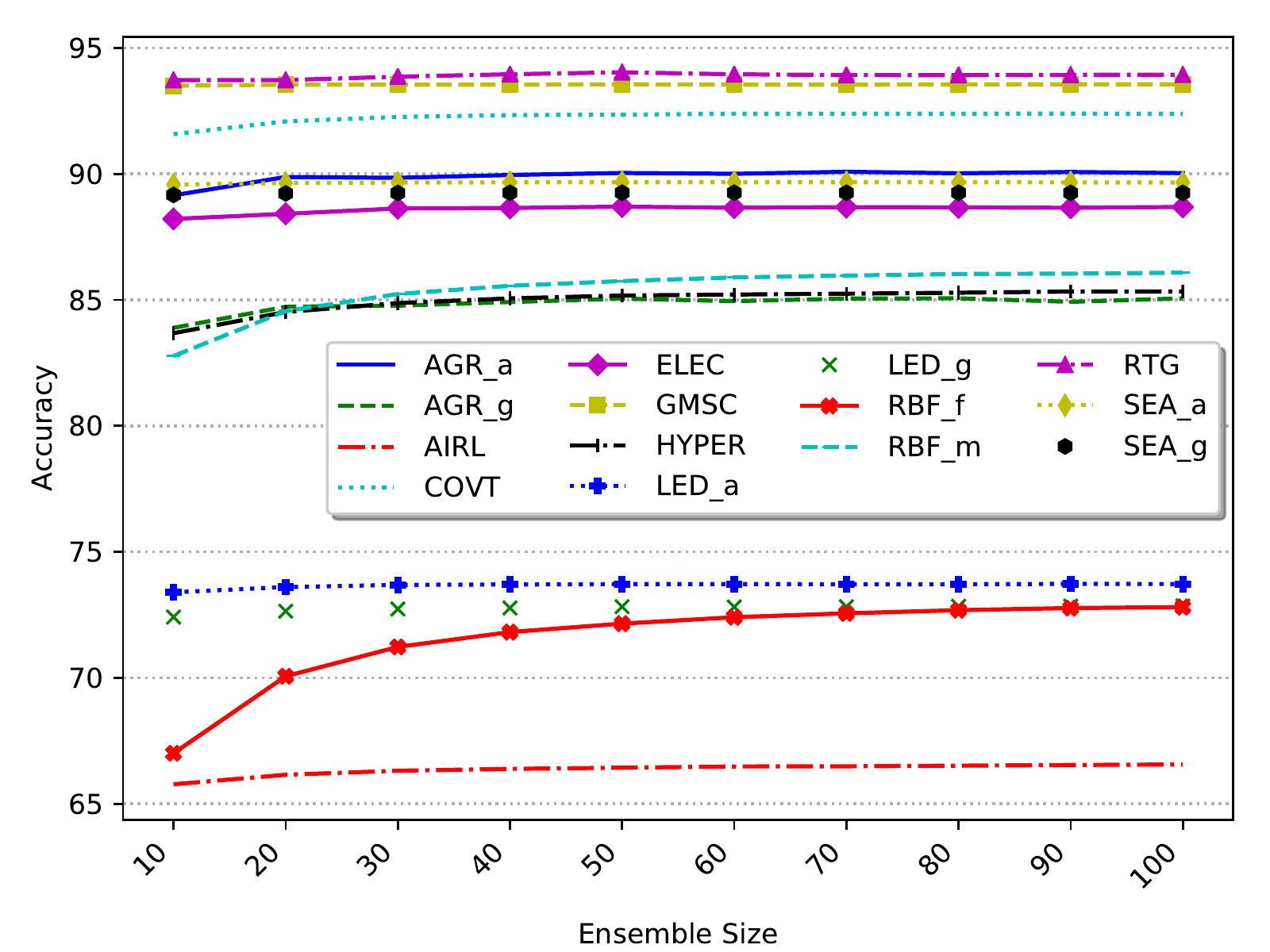}
\vspace{-0.25cm}
\caption{SRF accuracy evolution as the ensemble size increases}
\label{fig::srf_accuracy_evolution}
\end{figure}

Figure \ref{fig::srf_accuracy_evolution} shows the accuracy obtained by {\em SRF} when using a fixed number of learners in the $FS$. Comparing to Figure \ref{fig::arf_accuracy_evolution} one can appreciate that the accuracy of {\em SRF} converges faster than ARF100. With only 35 learners in $FS$ the differences in accuracy are less than 1 percentage point, while ARF100 required 50 learners to be within the same range.

Table \ref{tab:srf_vs_arf100} details the results for two {\em SRF} configurations (SRF F35 and F50) and ARF100. The $\Delta$ column always shows the difference with ARF100. When using 35 learners in the front set of {\em SRF}, the differences  observed in terms of accuracy are marginal in most of the tests, except for the two RBF datasets that are a little bit more noticeable; however, in terms of resources, {\em SRF} is only using one third of the trees that are used in ARF100. When using 50 learners, half of the trees compared to ARF100, {\em SRF} outperforms ARF100 in 11 of the 14 datasets, being the differences in the other 3 smaller than before. The reduction in size for F50 implies a speedup of 2.03x on average. Looking at the average for all datasets, we conclude that the average difference is very small (0.07 worse for F35 and 0.06 better for F50 wrt ARF100).  


\begin{table}[!h]
\centering
\footnotesize
\caption{Accuracy comparison between ARF100 and SRF with $|FS|=35$ and $|FS|=50$}
\label{tab:srf_vs_arf100}
\begin{tabular}{|r||c|cc|cc|}
\hline
& ARF100 & \multicolumn{2}{c|}{SRF F35}& \multicolumn{2}{c|}{SRF F50} \\
\hline
Dataset & Acc(\%) & Acc(\%) & $\Delta$ & Acc(\%) & $\Delta$ \\
\hline
AGR\_a & 89.73 & \textbf{90.13} & 0.40 &90.04 & 0.31 \\
AGR\_g & 84.53 & 84.81 & 0.28 &\textbf{85.05} & 0.52 \\
HYPER & 85.16 & 84.96 & -0.20 &\textbf{85.17} & 0.01 \\
LED\_a & \textbf{73.72} & 73.69 & -0.03 &\textbf{73.72} & 0.00 \\
LED\_g &\textbf{ 72.86} & 72.77 & -0.09 &72.82 & -0.04 \\
RBF\_f &\textbf{ 72.35} & 71.59 & -0.76 &72.15 & -0.20 \\
RBF\_m &\textbf{ 86.01} & 85.41 & -0.60 &85.75 & -0.26 \\
RTG & 93.91 & 93.82 & -0.09 &\textbf{94.03} & 0.12 \\
SEA\_a & 89.66 & 89.66 & 0.00 &\textbf{89.67} & 0.01 \\
SEA\_g & 89.24 & \textbf{89.25} & 0.01 &\textbf{89.25} & 0.01 \\
\hline
Average & 83.72 & 83.61 & -0.11 & 83.76 & 0.05 \\
\hline
\hline
AIRL & 66.25 & 66.34 & 0.09 &\textbf{66.43} & 0.18 \\
COVT & 92.31 & 92.30 & -0.01 &\textbf{92.35} & 0.04 \\
ELEC & 88.57 & 88.62 & 0.05 &\textbf{88.70} & 0.13 \\
GMSC & \textbf{93.55} & 93.54 & -0.01 &\textbf{93.55} & 0.00 \\
\hline
Average & 85.17 & 85.20 & 0.03 & 85.26 & 0.09 \\
\hline
\hline
Average & 84.13 & 84.06 & -0.07 & 84.19 & 0.06 \\
\hline
\end{tabular}
\end{table}

\subsection{\esrf}\label{sec::eval::esrf}

In this subsection we evaluate the complete {\em ESRF} ensemble, using both the swap and elastic component together. Once we have seen in the previous subsection how the swap component is able to significantly reduce the number of learners required in the ensemble, we want to see if the elastic component is able to dynamically determine the most appropriate size for the $FS$. As before, accuracy and ensemble size are compared against the ARF100 baseline.

\begin{figure}[!t]
\includegraphics[scale=0.55]{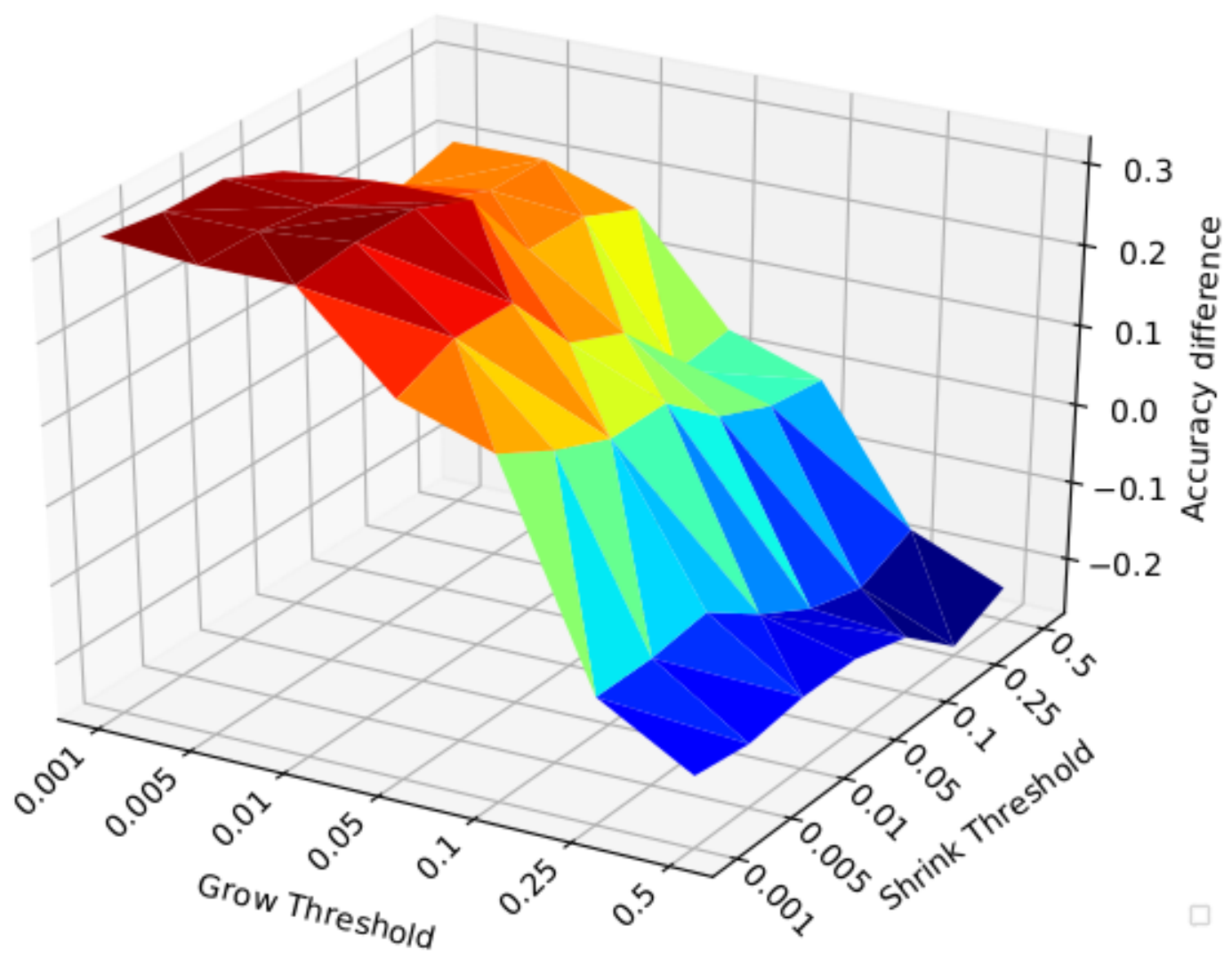}
\caption{Influence in the accuracy distance w.r.t. ARF100 of the grow and shrink thresholds for the AIRL dataset. Negative numbers mean ARF100 is better.}
\label{fig::esrf_sabana3d}
\end{figure}

In order to asses the impact of grow and shrink thresholds ($T_g$ and $T_s$ in Algorithm \ref{alg::esrf:check_if_resize}, respectively) in the accuracy and ensemble size, different sensitivity levels have been tested. Restrictive thresholds allow a resize operation only when the difference for $\Delta_{shrunk}$ or $\Delta_{grown}$ is in first decimal or higher; similarly, permissive thresholds allow differences in the second or third decimal. 
In general, more permissive thresholds make {\em ESRF} grow faster, which in time, reduces the differences with ARF100; however more restrictive thresholds have the opposite effect. Figure \ref{fig::esrf_sabana3d} shows an example of this trend for the AIRL dataset,  in which moving from more restrictive thresholds ($T_g=T_s=0.5$) to more permissive ones ($T_g=T_s=0.001$) improves the difference in accuracy from -0.2 to 0.3.  The detailed analysis for all datasets is not included due to the limitation in the number of pages. 

The two thresholds can also be set to target different target scenarios. For example, on hardware platforms, such as ARM boards typically used in Internet of Things (IoT) applications, it may be desirable to make the ensemble only grow if strictly needed and force it to reduce its size as soon as it can in order to save memory and computation time. This can be achieved by using a shrink threshold ($T_s$) lower than the grow threshold ($T_g$), for example $T_g=0.01$ and $T_s=0.001$. Table \ref{tab::esrf_tg0.1_ts0.01} details results obtained with this configuration. Observe that the average performance of {\em ESRF} and {\em ARF100} are very similar ({\em ESRF} worse with $-0.08$ difference). In 11 out of the 14 datasets, the difference in accuracy is never worse than -0.23, being able to outperform {\em ARF100}  by 1 percentage point in the AGR\_g dataset. The three datasets presenting more challenges to this configuration are RTG, RBF\_f, RBF\_m, being the last the worst performing one ($-0.81$ difference). As shown in the central and right part of this table, the main advantage of this configuration is that the average number of trees used in the ensemble is 22, (4.5 times less trees than {\em ARF100}, with the same proportional reduction in memory used); in 10 out of the 14 datasets {\em ESRF} never required more than 42 trees. This reduction in the ensemble size implies an average speedup of 3.84x, or in other words, may allow the use of devices up of 3.84x less powerful.

\begin{table*}[!t]
\centering
\footnotesize
\caption{ESRF comparison with ARF100. Resource-constrained scenario: $T_g=0.01$ and $T_s=0.001$}
\label{tab::esrf_tg0.1_ts0.01}
\begin{tabular}{|c||cc||ccc|cc|cccc|}
\hline
 & \multicolumn{2}{c||}{ARF100} & \multicolumn{3}{c|}{ESRF accuracy} & \multicolumn{2}{c|}{ESRF speedup} & \multicolumn{4}{c|}{ESRF size} \\ 
 \hline
Dataset & Acc(\%) & Time(s) & Acc(\%) & Rank &  Delta & Time(s) & Speedup & mean & stdev & max & min\\
\hline
AGR\_a & 89.73 & 22670.86 & \textbf{89.96} & 1 & 0.23 & 4423.66 & 5.12 & 11.86 & 2.49 & 25.00 & 10.00 \\
AGR\_g & 84.53 & 24630.29 &\textbf{ 85.55} & 1 & 1.02 & 4935.70 & 4.99 & 12.49 & 2.76 & 28.00 & 10.00 \\
HYPER & \textbf{85.16} & 20157.96 & 84.95 & 2 & -0.21 & 7506.14 & 2.69 & 32.96 & 8.07 & 50.00 & 10.00 \\
LED\_a & \textbf{73.72} & 15344.18 & 73.59 & 2 & -0.13 & 3439.01 & 4.46 & 19.74 & 6.32 & 38.00 & 10.00 \\
LED\_g & \textbf{72.86} & 15319.00 & 72.63 & 2 & -0.23 & 3439.02 & 4.45 & 19.39 & 6.35 & 40.00 & 10.00 \\
RBF\_f & \textbf{72.35 }& 19938.81 & 71.97 & 2 & -0.38 & 9385.82 & 2.12 & 44.88 & 9.04 & 57.00 & 10.00 \\
RBF\_m & \textbf{86.01} & 19638.30 & 85.20 & 2 & -0.81 & 6751.32 & 2.91 & 30.84 & 8.35 & 48.00 & 10.00 \\
RTG &\textbf{ 93.91} & 17669.59 & 93.45 & 2 & -0.46 & 7611.35 & 2.32 & 11.73 & 2.54 & 30.00 & 10.00 \\
SEA\_a &\textbf{ 89.66} & 14298.62 & 89.62 & 2 & -0.04 & 2913.33 & 4.91 & 14.72 & 1.63 & 20.00 & 12.00 \\
SEA\_g & \textbf{89.24 }& 13979.24 & 89.20 & 2 & -0.04 & 2902.59 & 4.82 & 14.72 & 1.63 & 20.00 & 12.00 \\
\hline
AIRL & 66.25 & 33588.54 & \textbf{66.34} & 1 & 0.09 & 12465.26 & 2.69 & 35.15 & 13.99 & 59.00 & 10.00 \\
COVT & \textbf{92.31} & 12625.36 & 92.14 & 2 & -0.17 & 3352.87 & 3.77 & 23.10 & 5.04 & 37.00 & 10.00 \\
ELEC & 88.57 & 843.06 &\textbf{ 88.66} & 1 & 0.09 & 234.45 & 3.60 & 23.03 & 5.61 & 42.00 & 10.00 \\
GMSC & \textbf{93.55 }& 2180.26 & 93.52 & 2 & -0.03 & 443.70 & 4.91 & 14.10 & 1.61 & 17.00 & 10.00 \\
\hline
Average & 84.13 & 16634.58 & 84.06 & 1.71 & -0.08 & 4986.02 & 3.84 & 22.05 & 5.39 & 36.50 & 10.29 \\
\hline
\end{tabular}
\end{table*}

\begin{table*}[!t]
\centering
\footnotesize
\caption{ESRF comparison with ARF100. $T_s=T_g=0.001$}
\label{tab::esrf_tg=ts_0.001}
\begin{tabular}{|c||cc||ccc|cc|cccc|}
\hline
 & \multicolumn{2}{c||}{ARF100} & \multicolumn{3}{c|}{ESRF accuracy} & \multicolumn{2}{c|}{ESRF speedup} & \multicolumn{4}{c|}{ESRF size} \\ 
 \hline
Dataset & Acc(\%) & Time(s) & Acc(\%) & Rank &  Delta & Time(s) & Speedup & mean & stdev & max & min\\
\hline
AGR\_a & 89.73 & 22670.86 & \textbf{90.47} & 1 & 0.74 & 4916.52 & 4.61 & 12.42 & 2.92 & 26.00 & 10.00 \\
AGR\_g & 84.53 & 24630.29 & \textbf{86.53} & 1 & 2.00 & 5348.60 & 4.60 & 13.31 & 4.16 & 38.00 & 10.00 \\
HYPER & 85.16 & 20157.96 & \textbf{85.38} & 1 & 0.22 & 14878.47 & 1.35 & 72.03 & 22.60 & 89.00 & 10.00 \\
LED\_a & \textbf{73.72} & 15344.18 & 73.67 & 2 & -0.05 & 3803.60 & 4.03 & 22.97 & 9.03 & 58.00 & 10.00 \\
LED\_g & \textbf{72.86} & 15319.00 & 72.74 & 2 & -0.12 & 4119.97 & 3.72 & 24.75 & 10.60 & 49.00 & 10.00 \\
RBF\_f & \textbf{72.35} & 19938.81 & 72.30 & 2 & -0.05 & 11706.91 & 1.70 & 57.24 & 17.10 & 89.00 & 10.00 \\
RBF\_m & \textbf{86.01} & 19638.30 & 85.87 & 2 & -0.14 & 12117.52 & 1.62 & 61.58 & 19.05 & 89.00 & 10.00 \\
RTG & \textbf{93.91} & 17669.59 & 93.61 & 2 & -0.30 & 6169.04 & 2.86 & 11.13 & 2.48 & 29.00 & 10.00 \\
SEA\_a & \textbf{89.66} & 14298.62 & 89.64 & 2 & -0.02 & 3665.54 & 3.90 & 21.38 & 6.05 & 38.00 & 10.00 \\
SEA\_g & \textbf{89.24 }& 13979.24 & 89.21 & 2 & -0.03 & 3467.54 & 4.03 & 19.14 & 5.29 & 35.00 & 10.00 \\
\hline
AIRL & 66.25 & 33588.54 & \textbf{66.57} & 1 & 0.32 & 20700.48 & 1.62 & 66.36 & 20.79 & 89.00 & 10.00 \\
COVT & 92.31 & 12625.36 & \textbf{92.34} & 1 & 0.03 & 5065.27 & 2.49 & 40.24 & 12.57 & 61.00 & 10.00 \\
ELEC & 88.57 & 843.06 & \textbf{88.70} & 1 & 0.13 & 254.51 & 3.31 & 26.92 & 7.23 & 43.00 & 10.00 \\
GMSC & \textbf{93.55} & 2180.26 & 93.53 & 2 & -0.02 & 521.10 & 4.18 & 18.30 & 4.44 & 35.00 & 10.00 \\
\hline
Average & 84.13 & 16634.58 & 84.33 & 1.57 & 0.19 & 6909.65 & 3.14 & 33.41 & 10.31 & 54.86 & 10.00 \\
\hline
\end{tabular}
\end{table*}

For other scenarios requiring higher accuracy, or in which memory or CPU time are not an issue, {\em ESRF} may use more sensitive thresholds in order to grow faster and reach higher accuracy requirements. For example, Table \ref{tab::esrf_tg=ts_0.001} shows the results that are obtained when using $T_g=T_s=0.001$. 
This configuration allows ESRF to grow larger in those datasets that were more challenging in Table \ref{tab::esrf_tg0.1_ts0.01}, while keeping similar average size for the rest of the datasets. This narrows the difference in accuracy to be not lower than  $-0.05$ in 12 out of the 14 datasets, and in the worst case $-0.3$ (RTG). In terms of number of learners and speedup the same table shows a reduction of the execution time by 3.14x  and the use of 33 learners on average, always compared to ARF100.

\section{Conclusions}\label{sec::conclusion}

This paper presents a new ensemble method for evolving data streams: \esrf ({\em ESRF}). {\em ESRF} aims at reducing the number of trees required by the reference \arf ({\em ARF}) ensemble while providing similar accuracy. {\em ESRF} extends {\em ARF} with two orthogonal components: 1) a swap component that splits learners into two sets based on their accuracy (only classifiers with the highest accuracy are used to make predictions); and 2) an elastic component for dynamically increasing or decreasing the number of classifiers in the ensemble. The experimental evaluation of {\em ESRF} and comparison with the original {\em ARF} shows how the two new components effectively contribute to reducing the number of classifiers up to one third while providing almost the same accuracy, resulting in speed-ups in terms of per-sample execution time close to 3x. In addition, a sensitivity analysis of the two thresholds determining the elastic nature of the ensemble has been performed, establishing a trade--off in terms of resources (memory and computational requirements) and accuracy (which in all cases is comparable to the accuracy achieved by ARF100).

{\em ESRF} has been implemented in the MOA (Massive Online Analysis) framework, an open source software environment for data stream mining, that implements a large number of data stream learning methods, including {\em ARF}. 


As part of our future work we plan to improve the resize logic, trying to make it more adaptive and correlated with the performance evolution of each tree and the drift detectors. 

\bibliographystyle{named}
\bibliography{mimic}

\end{document}